\documentclass[runningheads]{llncs}
\usepackage{graphicx}
\usepackage{hyperref}
\usepackage{caption}
\usepackage{subcaption}

\usepackage{xcolor}

\begin{document}
\title{Using Meta-learning to Recommend Process Discovery Methods\thanks{The authors would like to thank CNPq (National Council for the Scientific and Technological Development) for their financial support under Grant of Project 420562/2018-4 and 309863/2020-1 and the program “Piano di sostegno alla ricerca 2020” funded by Universit\`a degli Studi di Milano.}}
%
%
\author{Sylvio Barbon Junior\inst{1}\orcidID{0000-0002-4988-0702} \and
Paolo Ceravolo\inst{2}\orcidID{0000-0002-4519-0173} \and
Ernesto Damiani\inst{3}\orcidID{0000-0002-9557-6496} \and
Gabriel Marques Tavares\inst{2}\orcidID{0000-0002-2601-8108}}
\authorrunning{Barbon et al.}
%
\institute{Londrina State University (UEL), Londrina, Brazil\\
\email{barbon@uel.br} \and
Universit\`a degli Studi di Milano (UNIMI), Milan, Italy\\
\email{\{paolo.ceravolo, gabriel.tavares\}@unimi.it} \and
Khalifa University (KUST), Abu Dhabi, UAE\\
\email{ernesto.damiani@kustar.ac.ae}}
\maketitle              
\begin{abstract}
Process discovery methods have obtained remarkable achievements in Process Mining, delivering comprehensible process models to enhance management capabilities. However, selecting the suitable method for a specific event log highly relies on human expertise, hindering its broad application. Solutions based on Meta-learning (MtL) have been promising for creating systems with reduced human assistance. This paper presents a MtL solution for recommending process discovery methods that maximize model quality according to complementary dimensions. Thanks to our MtL pipeline, it was possible to recommend a discovery method with 92\% of accuracy using light-weight features that describe the event log. Our experimental analysis also provided significant insights on the importance of log features in generating recommendations, paving the way to a deeper understanding of the discovery algorithms. 

\keywords{Process discovery \and Meta-learning \and Model quality \and Recommendation.}
\end{abstract}
\section{Introduction}

According to a recent survey, process discovery is the most active research topic in Process Mining (PM)~\cite{GARCIA2019260}. 
High-quality process models can enhance management capabilities, revealing process deviations and improvement opportunities~\cite{kerremans2018market}. However, eliciting an appropriate process model may require significant efforts~\cite{van2010process}. One of the key design choices is selecting the proper discovery algorithm among several different options~\cite{1316839,tonbeta166,10.1007/978-3-642-38697-8_17,10.1007/978-3-319-06257-0_6,10.1007/978-3-319-19237-6_6,10.1007/978-3-319-07734-5_6}. Depending on the dimension to be optimized, the algorithms that offer the best performance, in terms of model quality, are different~\cite{Augusto2019TKDE}. Besides, the event log characteristics also impact the obtained model quality~\cite{Back2019,Augusto2019TKDE}. Matching a discovery algorithm to a specific business process is time-consuming and may often result in subjective and approximate decisions, requesting the intervention of human specialists~\cite{muarucster2006rule}.

Experiments with several different discovery algorithms showed a substantial performance complementarity since different algorithms perform best on different business process scenarios~\cite{Back2019,DEWEERDT2012654}. Multiple quality dimensions have been identified, among them, the literature has largely discussed \textit{fitness}, \textit{precision} and \textit{generalization}~\cite{augusto2019split}. However, other dimensions are also considered. For example, empirical studies have shown that block-structured process models are generally more understandable and less error-prone than unstructured ones. Structural complexity is quantified by the amount of branching in a process model. However, the structuredness of a model does not necessarily result in less understandability. If the structure is hierarchical, i.e., the flow can be divided recursively into parts having a single entry and exit points, its understandability increases~\cite{mendling2008metrics}. On the other side, a model has to reflect the complexity intrinsically expressed in the event logs it has to represent. Log variability, expressed in terms of entropy measures, has been proposed as an indicator of the process discovery paradigm to be preferred~\cite{Back2019}. Log complexity is identified as a key feature affecting the performance of discovery algorithms~\cite{Augusto2019TKDE}.
In other words, it is important to choose an algorithm considering the features characterizing the analyzed event log. However, these characteristics cannot be straightforward abstracted, besides linking them to a particular algorithm may be very tricky. PM specialists, thanks to their professional experience, developed the knowledge to appropriately sort out discovery algorithms but their work is time-consuming and costly. 

To overcome these issues, in this work, we study a method to \textit{automate the selection of the optimal process discovery algorithm} given an event log. 
Techniques to enhance automated process discovery have been proposed in~\cite{Augusto2019}. Using meta-heuristic optimization, the authors perturb the directly-followed graph, an inner data representation of many discovery algorithms, to boost the discovered model’s accuracy. However, the approach is not algorithm agnostic, limiting the applicability in more scenarios. Another consideration is that the optimization is focused either on \textit{precision} or \textit{fitness}, ignoring other quality criteria. 

Meta-learning (MtL) has achieved considerable success in emulating expert decisions in the form of a recommender system~\cite{aguiar2019meta,oyamada2020towards,he2021automl}. MtL is a learning procedure applied to the metadata describing other learning procedures to identify the metadata set that brings the best performance. 
With MtL high potential in mind, we developed a data-driven approach to recommend a suitable process discovery algorithm according to PM quality metrics.
Our MtL approach provides a novel tool to identify a discovery technique that balances model quality metrics given the specific features characterizing an event log. Moreover, it provides a method to exhaustively compare discovery algorithms in terms of quality metrics concerning event logs features and, more generally, to study the relationship between quality results and event logs features. 

More specifically, the paper is organized as follows. Section~\ref{mm} delves into the details of the proposed MtL framework, introduces the required theoretical foundations, exposes the feature extraction step and the materials used for experimentation. Section~\ref{rd} reports the results from the experiments and presents a discussion of the relationship between process descriptors, discovery techniques, and model quality. Finally, Section~\ref{cc} summarizes and concludes the paper.

\section{Methodology}~\label{mm}

In this section, we present the procedure executed to study the applicability of MtL to the selection of process discovery algorithms. The discussion focuses on the experimental design. The reader interested in the basic notions of process discovery is referred to~\cite{Aalst16}. For an overview on feature extraction with event logs, we recommend~\cite{Back2019,10.1007/978-3-030-70650-0_11}.
The material used in the experiments, along with event logs and implementation, is publicly available\footnote{\url{https://github.com/gbrltv/process_meta_learning}}.

\subsection{Proposed approach}

\begin{figure}[ht!]
    \centering
    \includegraphics[width=\textwidth]{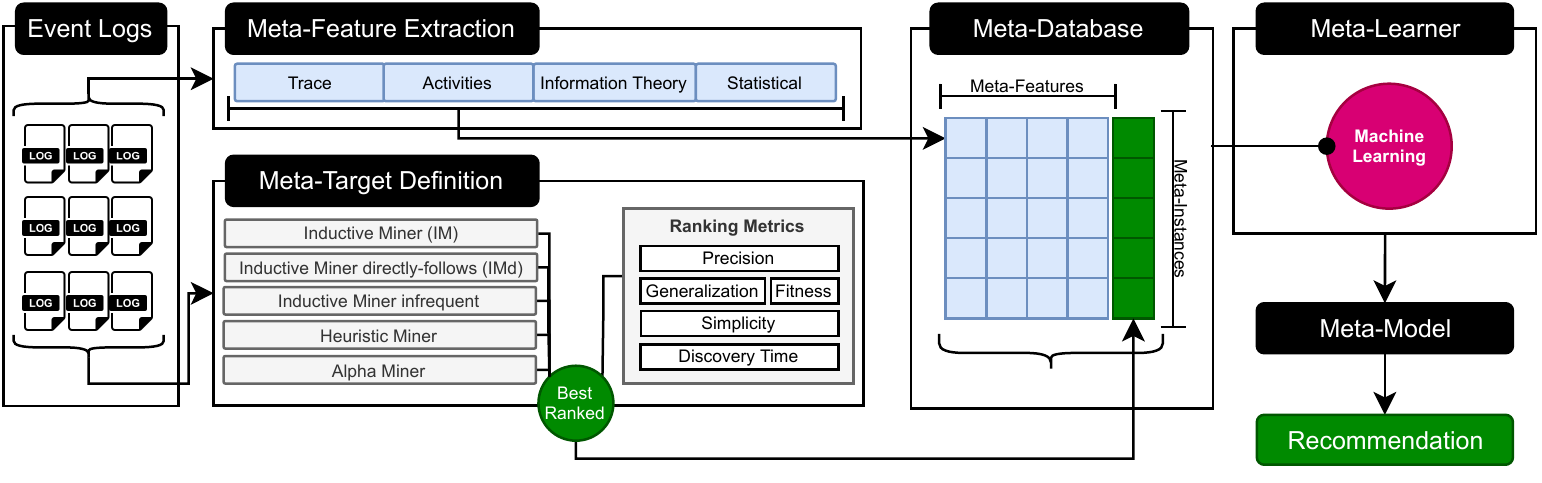}
    \caption{Overview of the proposed MtL approach}
    \label{overview}
\end{figure}

We created an MtL procedure grounded on rich PM meta-features for suggesting the best-ranked discovery algorithm. The ranking follows process model quality criteria and computational time for discovery. The workflow implementing our MtL approach is composed of five steps. Figure~\ref{overview} presents them as follows. \textit{Meta-Feature Extraction} is a step devoted to describing the event log characteristics.
\textit{Meta-Target Definition} identifies the discovery algorithms to be assessed using quality metrics. 

\begin{figure}[ht!]
    \centering
    \includegraphics[width=0.4\textwidth]{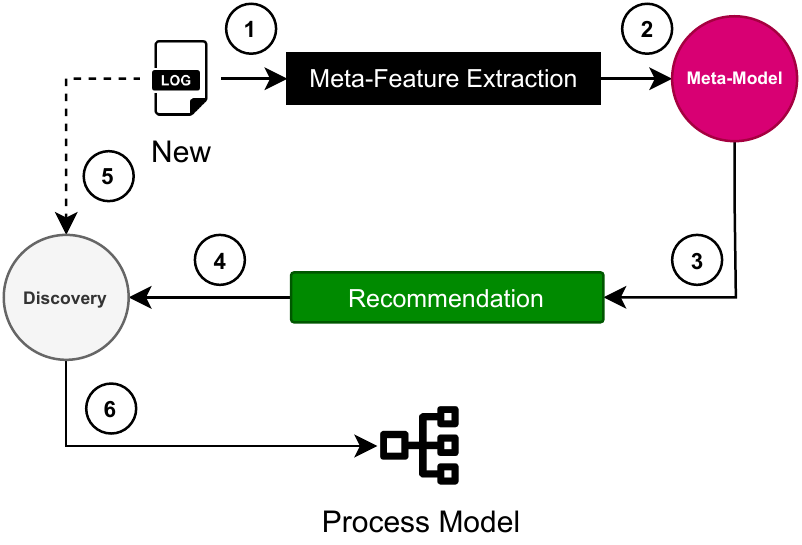}
    \caption{Application of our MtL approach, 1) raw event log has its features extracted, 2) features are processed using the meta-model, 3) the recommendation is outputted, 4) using the suggested algorithm, processes the new event log (5) to extract the process model (6).}
    \label{application}
\end{figure}

\textit{Meta-Database}, through this step, the meta-features are combined, forming meta-instances used to train the meta-model. \textit{Meta-Learner}, is a step grounded on machine learning for inducing the meta-model using the obtained meta-instances. \textit{Meta-Model}, is the final model able to output the recommendation of a process model algorithm. When recommending a process discovery method for a new event log, meta-features of this new resource are extracted and forwarded to the created meta-model. The recommended discovery algorithm can be executed on the event log generating a process model, as in Figure~\ref{application}.

\subsection{Event logs as meta-instances}

Event logs play a significant role in our approach because they are the search space representation towards creating our meta-database. After a given event log has extracted its meta-features (Section \ref{mtl_extraction}) and labelled by a meta-target (Section \ref{algo}), it forms a meta-instance. Thus, we aim at building a highly heterogeneous set of logs capable of representing a wide range of possible business process behaviors. Hence, the relationship between business process characteristics and quality metrics can be better represented. The data selected contains both real and synthetic event logs.

Regarding synthetic event logs, the first set comes from the Process Discovery Contest (PDC) 2020\footnote{\url{https://www.tf-pm.org/competitions-awards/discovery-contest}}. PDC is an annually organized contest that aims to evaluate the efficiency of process discovery techniques. Naturally, the PDC 2020 dataset explores a broad range of characteristics distributed in 192 logs. The main behaviors are dependent tasks, loops, \textit{or} constructs, invisible and duplicate tasks, and noise. Moreover, we extracted 750 event streams from~\cite{9124702}. These logs were built in the context of online PM, their main goal is depicting drifting scenarios, i.e., where a change in the behavior occurs while the process is acting. The streams were generated from one model perturbed by 16 change patterns, which are operations applied to the model, such as adding, removing, looping, swapping, or parallelizing fragments. Additionally, the event logs are arranged based on four drift types, five noise percentages, and three trace lengths. The last group of synthetic logs was presented in the context of evaluating encoding capabilities in event logs~\cite{10.1007/978-3-030-70650-0_11}. Five process models were simulated to generate 140 event logs containing six anomaly types with four incidence percentages.

The real-life event logs set contains six logs from the Business Process Intelligence Challenges (BPIC)\footnote{\url{https://www.tf-pm.org/resources/logs}}, the environmental permit\footnote{\url{https://doi.org/10.4121/uuid:26aba40d-8b2d-435b-b5af-6d4bfbd7a270}}, helpdesk\footnote{\url{https://doi.org/10.17632/39bp3vv62t.1}} and sepsis\footnote{\url{https://doi.org/10.4121/uuid:915d2bfb-7e84-49ad-a286-dc35f063a460}} logs. The final dataset contains 1091 event logs, their main characteristics are exposed in Table~\ref{tab:log_stats}. As demonstrated in the table, the event logs contain a wide range of behaviors, with a diverse number of cases, events and activities to cover a variety of meta-instances patterns.

\begin{table}[!ht]
\centering
\scalebox{0.91}{
\begin{tabular}{l|llllll}
\hline
Name & \#Logs & \#Cases & \#Events & \#Activities & Trace length & \#Variants\\
\hline
PDC 2020 & 192 & 1k & 6.7k-66k & 14-36 & 3-300 & 503-1k\\
Streams & 750 & 100-1k & 900-13.5k & 15-16 & 2-83 & 22-204\\
Encoding & 140 & 1k & 10k-44k & 22-406 & 1-50 & 383-1k\\
\hline
BPIC12 & 4 & 5k-13k & 31k-262k & 7-24 & 2-175 & 17-4.3k\\
BPIC13 & 2 & 1k-7k & 6k-65k & 4 & 1-123 & 183-1.5k\\
Env. permit & 1 & 1.4k & 8.5k & 27 & 1-25 & 116\\
Helpdesk & 1 & 4.5k & 21k & 14 & 2-15 & 226\\
Sepsis & 1 & 1k & 15k & 16 & 3-185 & 841\\
\hline
\end{tabular}
}
\caption{Detailed event logs statistics}
\label{tab:log_stats}
\end{table}

\subsection{Meta-Feature extraction}
\label{mtl_extraction}

A challenge in our approach is to correctly capture log behavior using an abstract and representative set of descriptors. Moreover, it is worth noting that the feature extraction operation should have a low computational cost; otherwise, extracting meta-features would be more costly than testing the target event log on several process discovery techniques. Thus, the goal is to propose a lightweight group of features with good representational capability. 

For trace level descriptors, trace lengths and variants were used as light-weight indicators of process complexity. Regarding trace lengths, we extracted 29 descriptors: minimum, maximum, mean, median, mode, standard deviation, variance, the 25th and 75th percentile of data, interquartile range, geometric mean and standard variation, harmonic mean, coefficient of variation, entropy, and a histogram of 10 bins along with its skewness and kurtosis coefficients. In counting the number of trace variants, we extracted additional 11 features: mean, standard variation, skewness coefficient, kurtosis coefficient, the ratio of the most common variant to the number of traces, and ratios of the top 1\%, 5\%, 10\%, 20\%, 50\% and 75\% to the total number of traces. To capture features at the activity level, we selected three groups: all activities, start activities, and end activities. For each group, we extracted a set of 12 descriptors: number of activities, minimum, maximum, mean, median, standard deviation, variance, the 25th and 75th percentile of data, interquartile range, skewness, and kurtosis coefficients. 

For log level descriptors, we obtained the number of traces, unique traces, and their ratio, along with the number of events. Recently, entropy measures were proposed in the context of business processes to capture log variability~\cite{Back2019}. In~\cite{Back2019}, the authors propose several entropy measures to identify whether the log better fits declarative or imperative mining. Consequently, these metrics measure log structuredness, a very relevant descriptor of log complexity. The entropy measures are further divided into three groups, in-trace frequency, language-inspired, dynamic system, and molecular structural analysis. This way, we extracted 14 entropy measures: trace, prefix, \textit{k}-block difference and ratio (applied with \textit{k} values of 1, 3 and 5), global block, \textit{k}-nearest neighbor (applied with \textit{k} values of 3, 5 and 7), Lempel-Ziv, and Kozachenko-Leonenko.

In total, the 93 features we extracted cover several complementary perspectives, such as statistical dispersion, the probability distribution shape, and tendency, log structuredness and variability, among others. 

\subsection{Meta-targets to be recommended}\label{algo}

We selected five meta-target candidates considering their wide historical use in PM:
$\alpha$-Miner (AM), Heuristic Miner (HM), Inductive Miner (IM), Inductive Miner--infrequent (IMf), and Inductive Miner--directly-follows (IMd).
All selected algorithms are suitable meta-targets in our experiment as they share the same input and output objects. They require an event log as input and express the discovered process model using the Petri net notation.

AM is one of the first approaches in PM that deals with concurrency. It is considered a baseline process discovery algorithm~\cite{1316839} as many subsequent ideas were developed from it. The method's first step is to scan the log looking for relevant patterns, known as directly-follows relations. These relations define, for instance, if an activity is always or never followed by another. This way, basic control-flow patterns can be derived from the relations of all activities in the log, creating a log footprint. A process model is then generated by aggregating basic patterns in more abstract ones. Albeit AM's historical relevance, there are many known limitations such as the lack of support for loops, weakness against noise, and no consideration of frequencies~\cite{Aalst16}.

In an effort to include frequencies into account, HM was introduced, applying the idea that infrequent transitions should not be represented in the model~\cite{tonbeta166}. HM uses causal nets, a modeling notation able to incorporate activities and causal dependencies. The first step HM performs is to compute frequencies of the directly-followed relation for all activities. Using frequencies, a dependence measure is calculated and further used to guide creating a dependency graph (using heuristics). Arcs not conforming to a threshold frequency are removed from the dependency graph. In a final step, splits and joins are introduced to represent concurrency.

The IM family of algorithms approaches the discovery problem from a divide-and-conquer perspective, recursively splitting the event log into sub-logs~\cite{10.1007/978-3-642-38697-8_17}. First, IM builds a directly-followed graph. Then, based on process tree operators (exclusive-choice cut, sequence cut, parallel cut, and redo-loop cut), the log is split into sub-logs. The cutting procedure is repeated until a sub-log with only one activity is reached. The sequence of operators and activities can then be represented as a process tree, easily converted to other process modeling notations. A crucial point of IM is that it produces a sound process model capable of replaying the complete event log. That is, fitness is guaranteed. However, due to no support for duplicate or silent activities when building the process tree, IM may produce underfitting models~\cite{Aalst16}. Finally, IM is incapable of handling fixed-length repetitions and cannot manage infrequent behavior.

The basic IM algorithm is highly extendable. Thus, many variants have been proposed~\cite{10.1007/978-3-319-06257-0_6,10.1007/978-3-319-19237-6_6,10.1007/978-3-319-07734-5_6}. IMf incorporates the eventually-follows relation to better deal with incompleteness in event logs~\cite{10.1007/978-3-319-06257-0_6}. Moreover, IMf introduces activity and arc filters to remove infrequent behavior and produce a more precise model. As a consequence, the fitness guarantee does not hold. A drawback of both IM and IMf is scalability due to the recursive split, which requires multi-pass analysis. IMd was proposed to overcome this problem. For that, IMd performs the recursive step only on the directly-follows graph, without partitioning sub-logs. This improvement is especially applicable for scenarios where huge logs should be mined. However, the non-partitioning adaptation hurts the rediscoverability property. Hence, perfect fitness is not preserved. Similar to IM, IMd also cannot handle duplicate and silent activities.

\subsection{Meta-Database for Meta-Model creation}

Our meta-database regards a suitable matching of meta-instance and meta-target. The definition of the best choice of meta-target for a given event log is based on a ranking strategy considering several complementary perspectives, i.e., there is no unique metric that captures the overall model quality. 
In this work, we adopt the four traditional model quality metrics used in PM: \textit{fitness}, \textit{precision}, \textit{generalization} and \textit{simplicity}.

The \textit{fitness} metric aims to measure how much behavior in the log is allowed by the model~\cite{Buijs2014}, that is, to which extent traces compare valid execution paths derived from the model~\cite{ROZINAT200864}. It is measured by applying replay techniques, which aim to compare log and model. Hence, an excellent fitness value is achieved when the model can replay all traces in the log. Moreover, fitness can be measured at different levels, e.g., event, case, and log, and different measures have been proposed in the literature~\cite{ROZINAT200864,de2007genetic,aalstreplaying,BertiA19}. In this work, we adopt the fitness proposed in~\cite{BertiA19} due to its improvement over previous versions, namely, scalability and avoidance of known problems such as token flooding. In~\cite{BertiA19}, the authors use a token-based replay technique that, along with a list of transitions enabled during the replay, produces the number of (i) consumed tokens ($c$), (ii) produced tokens ($p$), (iii) missing tokens ($m$), and (iv) remaining tokens ($r$). Assuming each case $L_i$ of the log $L$, the log level fitness value ($f$) is calculated according to Equation~\ref{eq:fit}.

\begin{equation}\label{eq:fit}
    f = \frac{1}{2}\Bigg(1-\frac{\sum_{L_i \in L}^{} m_i}{\sum_{L_i \in L}^{} c_i}\Bigg) + \frac{1}{2}\Bigg(1-\frac{\sum_{L_i \in L}^{} r_i}{\sum_{L_i \in L}^{} p_i}\Bigg)
\end{equation}

The following quality dimension, \textit{precision}, measures the extent of the model allowed behavior that is not observed in the log~\cite{Buijs2014}. A poor precision indicates an underfitting model, i.e., it allows too many patterns not present in the event log. As precision measures negative examples, i.e., behavior allowed by the model but not seen in the log, the calculation is complex and often depends on approximation, for instance, when there are loops. In our experiment, we adopt the precision proposed in~\cite{10.1007/978-3-642-15618-2_16} because it is more efficient and granular than previous measures. The authors introduce the concept of escaping edges, which are the borders between the log’s behavior and the model where the model deviates from available observations in the log. The precision is derived from the quantification of escaping edges and their frequencies. The more the model differs from the log, the less precise it is. Formally, assume $L$ as the log, $M$ as the Petri net model, $E_E$ as model escaping edges, $A_T$ as model allowed tasks, $\sigma$ as a trace, and that state $s_j^i$ denotes the j--th state of $\sigma_i$, the precision ($p$) is calculated according to Equation~\ref{eq:prec}.

\begin{equation}\label{eq:prec}
    p = 1 - \frac{\sum_{i=1}^{|L|}\sum_{j=1}^{|\sigma_i|+1} |E_E(s_j^i)|}{\sum_{i=1}^{|L|}\sum_{j=1}^{|\sigma_i|+1} |A_T(s_j^i)|}
\end{equation}

Precision measures how underfitting a model is; contrarily, \textit{generalization} goes in the opposite direction by measuring model overfitting. Event logs present a sample of possible behavior allowed by the system, this implies that there may be valid execution traces not present in the log because they were not executed yet. Process models should then describe the log behavior but also generalize it to some extent~\cite{Aalst16,Buijs2014}.

This way, measuring generalization is challenging because the goal is to estimate how well the model describes an unknown system. A practical way of achieving a value is to compute how often subsets of the model are visited when replaying the event log. Hence, a good generalization means all model subsets are frequently visited, whereas a bad generalization means some parts of the model are rarely used. We adopt the computation proposed in~\cite{Buijs2014} as it covers the generalization based on model usefulness. Equation~\ref{eq:gen} shows how generalization ($g$) is computed.

\begin{equation}\label{eq:gen}
    g = 1 - \frac{\sum_{i=1}^{\#nodes}(\sqrt{\#executions_{node_i}})^{-1}}{\#nodes}
\end{equation}

\textit{Simplicity} is the last metric used in this work to evaluate model quality. Inspired by Occam's Razor principle, ``one should not increase, beyond what is necessary, the number of entities required to explain anything", the idea behind the simplicity is to indicate how complex is a model~\cite{Aalst16,Buijs2014}. That is, the more simple the model structure that reflects log behavior, the better its intelligibility. We adopt the simplicity measure proposed in~\cite{VAZQUEZBARREIROS2015315}, which is based on the weighted average degree for a place/transition in the Petri net, ultimately defined by the sum of input and output arcs. Assume $S$ as the sum of input and output arcs for all places and transitions in the Petri net, Equation~\ref{eq:simp} defines how simplicity ($s$) is calculated.

\begin{equation}\label{eq:simp}
    s = \frac{1}{1 + max(0, mean(S) - 2)}
\end{equation}

The quality metrics capture complementary views of the process model. This way, a discovery technique needs to provide a balance within the metrics. Otherwise, not considering one or more dimensions may lead to poor models, as shown in~\cite{Buijs2014}. Considering the importance of balancing perspectives to achieve an adequate model, we propose a ranking mechanism that aggregates all dimensions. 

Table~\ref{tab:ranking} provides an example of the ranking to identify the best discovery technique for a single event log. The log $L$ is submitted to three discovery algorithms ($A_1$, $A_2$, $A_3$), then the four quality metrics are extracted using the discovered model. Following, a rank is built for each metric considering the performance of the algorithms. A final rank ($R$) is produced by averaging the metrics ranks. In our example, $R(A_1)$ is 1.75, $R(A_2)$ is 1.5, and $R(A_3)$ is 2.75. Thus, we can conclude that $A_2$ is the discovery algorithm that produces the best model for log $L$ because it has the lowest $R$ value. Hence, when building the meta-database, $A_2$ is the meta-target for log $L$. Alternative quality metrics could be exploited to deliver specific recommendations. To exemplify this possibility, in our work, we also exploit the computation time as a possible metric when recommending a discovery technique focusing on fast processing.

In the final step of our experiment, a machine learning algorithm, here called meta-learner, is employed to induce a meta-model using the meta-database. Once a meta-model has been created, any event log can be linked to the recommended discovery algorithm by extracting its meta-features and running the meta-model.

\begin{table}[!ht]
\centering
\begin{tabular}{lllllllllll}
\hline
Log & Algorithm & $f$ & $p$ & $g$ & $s$ & $R_f$ & $R_p$ & $R_g$ & $R_s$ & $R$\\
\hline
$L$ & $A_1$ & 1 & 0.27 & 0.91 & 0.6 & 1 & 2 & 2 & 2 & 1.75\\
$L$ & $A_2$ & 0.98 & 0.38 & 0.93 & 0.64 & 3 & 1 & 1 & 1 & 1.5\\
$L$ & $A_3$ & 0.99 & 0.2 & 0.9 & 0.59 & 2 & 3 & 3 & 3 & 2.75\\
\hline
\end{tabular}
\caption{Example of ranking process discovery algorithms. The final rank ($R$) is generated from the average rank of each metric. In this example, $A_2$ is the recommended discovery technique for log $L$ as it maximizes the four process model quality metrics, i.e., produces the lowest $R$ value.}
\label{tab:ranking}
\end{table}

\section{Results and Discussion}~\label{rd}

In this section, we, first, report an experiment that compares the discovery techniques presented in~\ref{algo} using quality metrics and discovery time. After, we focus on a restricted set of techniques, namely AM, HM and IM, to assess the stability of the experimental results.

\subsection{Time as a requirement}

In addition to the four quality criteria, we included the algorithms’ time to discover the process model. Though it is not directly related to model quality, the discovery time cost can be a constraint in real scenarios, so it can be used to compose the final ranking in the meta-database.

\begin{figure}[!ht]
    \centering
    \begin{subfigure}[b]{0.47\textwidth}    
       \includegraphics[width=1\textwidth]{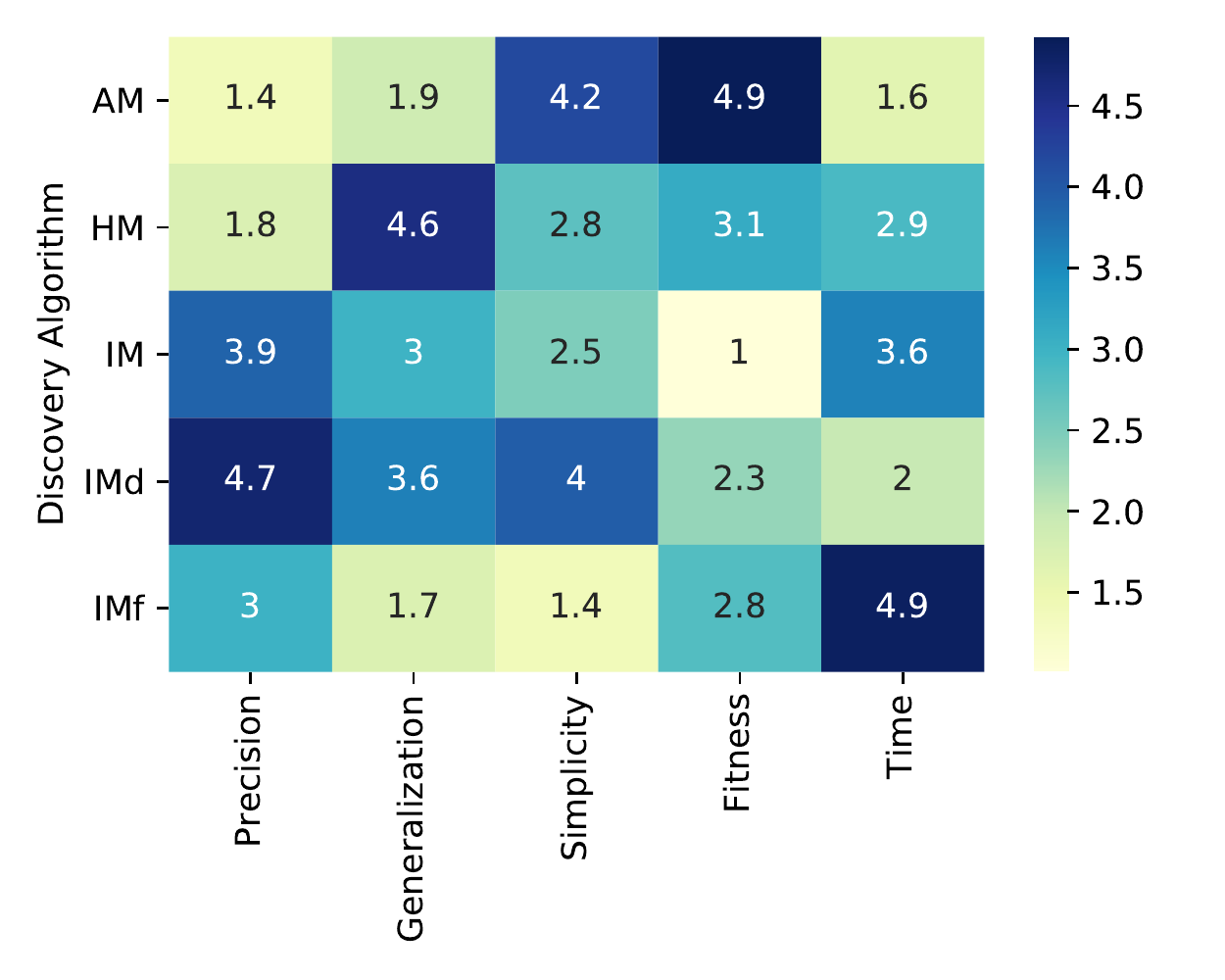}
    \caption{Metrics ranking}
    \label{rank5}
    \end{subfigure}
    \hfill
    \begin{subfigure}[b]{0.52\textwidth}
        \includegraphics[width=1\textwidth]{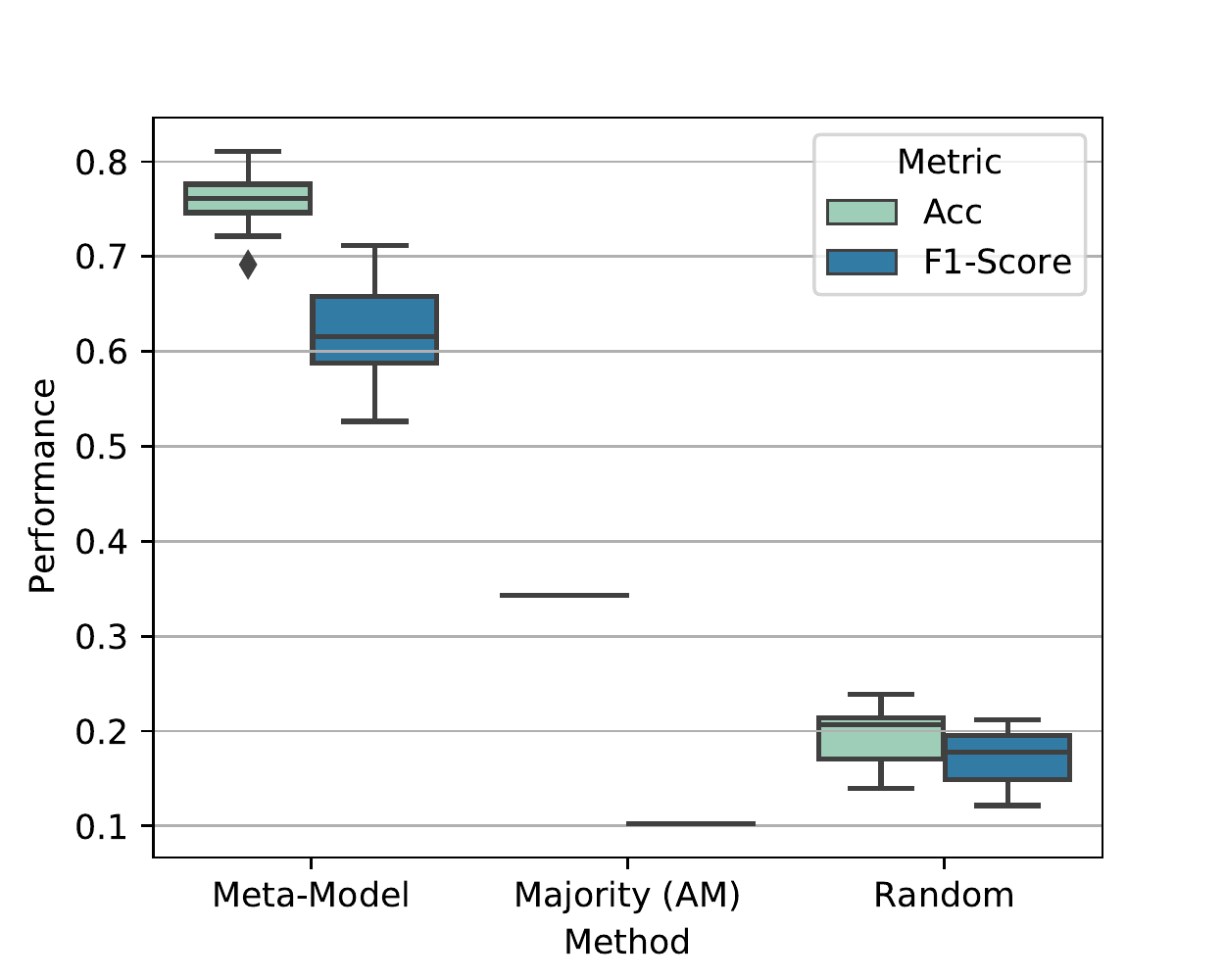}
    \caption{Recommendation performance}
    \label{rank5_perf}
    \end{subfigure}
    \caption{Discovery algorithm rankings across all event logs considering each dimension. Given event log features, the methods indicate the discovery algorithm that maximizes model quality. This experiment compares five discovery algorithms using five quality dimensions (fitness, precision, generalization, simplicity and time).}
\end{figure}

Figure~\ref{rank5} reports the discovery algorithm's average position across all event logs considering the four quality criteria and discovery time. These results already confirm theoretical findings. For instance, IM's \textit{fitness} takes the three first positions. Both IMd and IMf do not guarantee perfect fitness; still, they perform better than AM and HM. IMd averages 2.3 while IMf 2.8. Considering that IMd does not partition sub-logs, such an important characteristic of its predecessor, the fitness result is good. On the other hand, IMf, which incorporates the eventually-follows relation, is worse than IMd from a fitness perspective. HM and AM follow with 3.1 and 4.9 average positions. AM is by far the worst-performing algorithm for fitness purposes, a problem recognized since its inception. The inability to deal with incompleteness and weakness against noise has a high toll on fitness performance. Another explanation for the IM family overcoming HM and AM is that IM algorithms produce a sound model while HM and AM do not have this guarantee. In fact, unsound models tend to produce low fitness values.

Positions change considerably when evaluating \textit{precision}. AM becomes the best performing algorithm (1.4), followed by HM (1.8). IMf, IM, and IMd come next, averaging 3, 3.9, and 4.7, respectively. The IM family performs poorly in precision due to its extensive use of hidden transitions, which contribute to creating underfitting models, i.e., models that allow much behavior outside the event log. Contrarily, AM and HM hold a superior control of escaping edges, strengthening precision values. From the generalization perspective, IMf is the best algorithm (1.7), followed closely by AM (1.9). These algorithms excel in generating balanced models where different regions are similarly visited during trace replaying. Particularly, IMf's ability to remove infrequent behavior reflects in good generalization values. Indeed, including low-frequency traces into the model hurts generalization capabilities. IM, IMd, and HM averaging 3, 3.6, and 4.6 are the worst-performing discovery algorithms in this dimension. 

\textit{Simplicity} is an index depending uniquely on the model (other metrics require the event log), it rewards concise models. Again, IMf is the best performing method (1.4) thanks to its cleaning process, removing infrequent behavior. IM and HM follow averaging 2.5 and 2.8, and IMd and AM produce the most complex models, averaging 4 and 4.2, respectively. 

\textit{Time} analysis tends to benefit simpler algorithms as they require fewer steps to generate a model. This explains why AM has the best time performance, averaging 1.6. IMd comes after (2) since its primary design purpose is focused on time improvement. HM, a more robust algorithm, comes third with 2.9 as average. Lastly, IM and IMf average at 3.6 and 4.9. These two algorithms are the slowest as they demand a multipass recursive analysis of the event log. Especially, IMf spends more time to produce the model as it has additional steps to remove infrequent behavior.

In the next experiment, we evaluate the meta-model's efficiency to indicate the best discovery method for a given event log.
For this experiment, we used a Random Forest~\cite{breiman2001random} classifier due to its stability and wide use in similar researches. A holdout strategy was applied to divide the meta-database into train and test sets, using a 75\%/25\% split. We compared three methods: our MtL, majority voting, and random selection. Not having other literature references, we used majority voting and random selection as baseline approaches, used for comparison reasons. In generating the meta-model, we performed 30 experiments to reduce the influence of outliers. Figure~\ref{rank5_perf} presents the classification results, which were measured using the accuracy and F-score metrics. The most frequently indicated algorithm is AM, then the majority method always applies AM. The performance clearly shows that our meta-model overcomes the baseline methods. Our approach demonstrates a viable solution for recommending discovery algorithms with an accuracy of 0.76 and an F-score of 0.62. The majority method has the next best accuracy (0.34), followed by the random method (0.19). Regarding F-score, random selection reaches 0.17 while majority voting stays at 0.1. The advantage of applying MtL in comparison to unintelligent approaches is clear. Moreover, the experiment confirms a relationship between event log characteristics and model quality depending on the discovery technique. According to log features, there is an algorithm that maximizes the quality of the discovered model.

\begin{figure}[ht!]
    \centering
    \begin{subfigure}[b]{0.49\textwidth}    
        \includegraphics[width=0.85\textwidth]{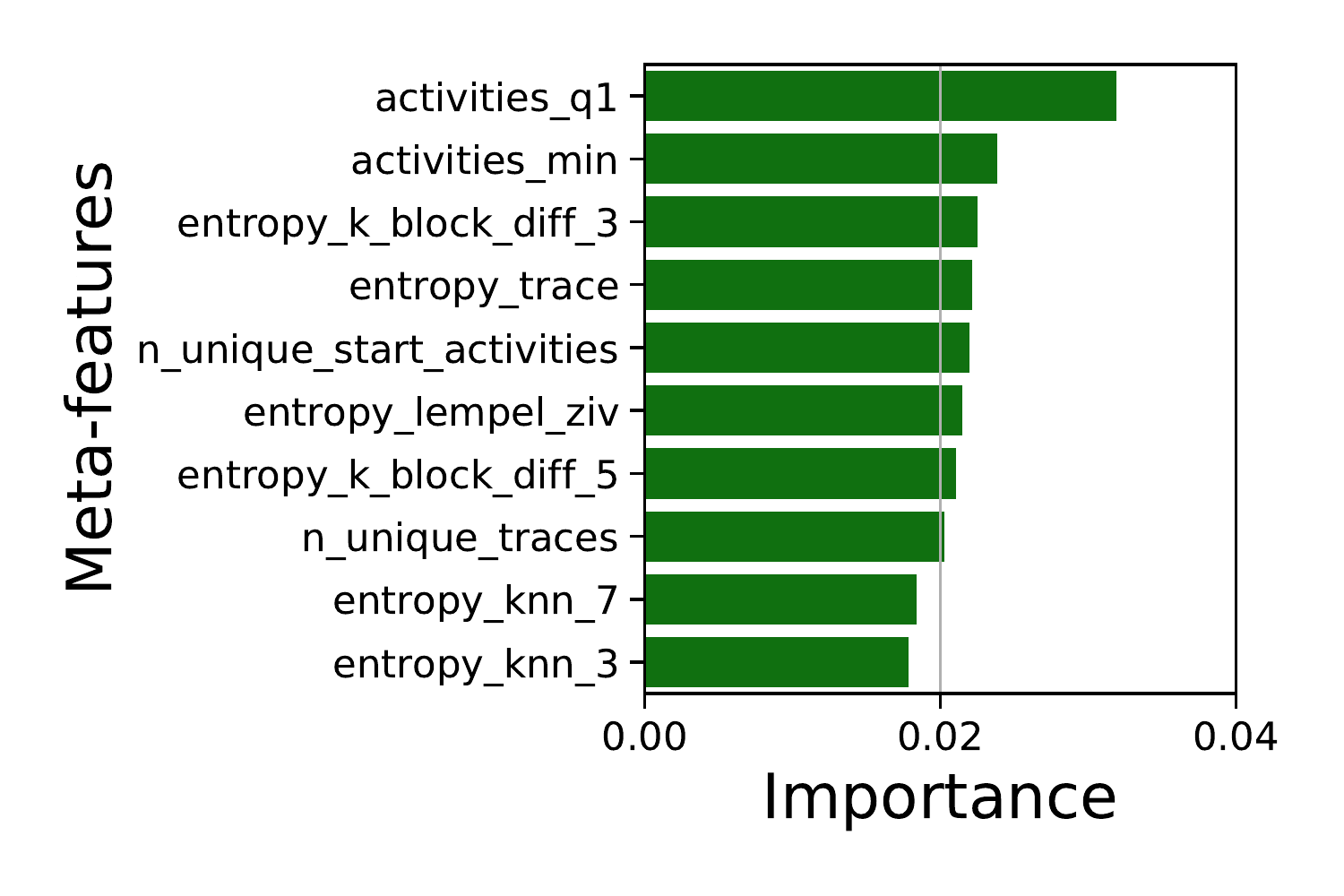}
        \caption{Top 10 features}
        \label{rank5_top10_features}
    \end{subfigure}
    \hfill
    \begin{subfigure}[b]{0.5\textwidth}
        \includegraphics[width=0.85\textwidth]{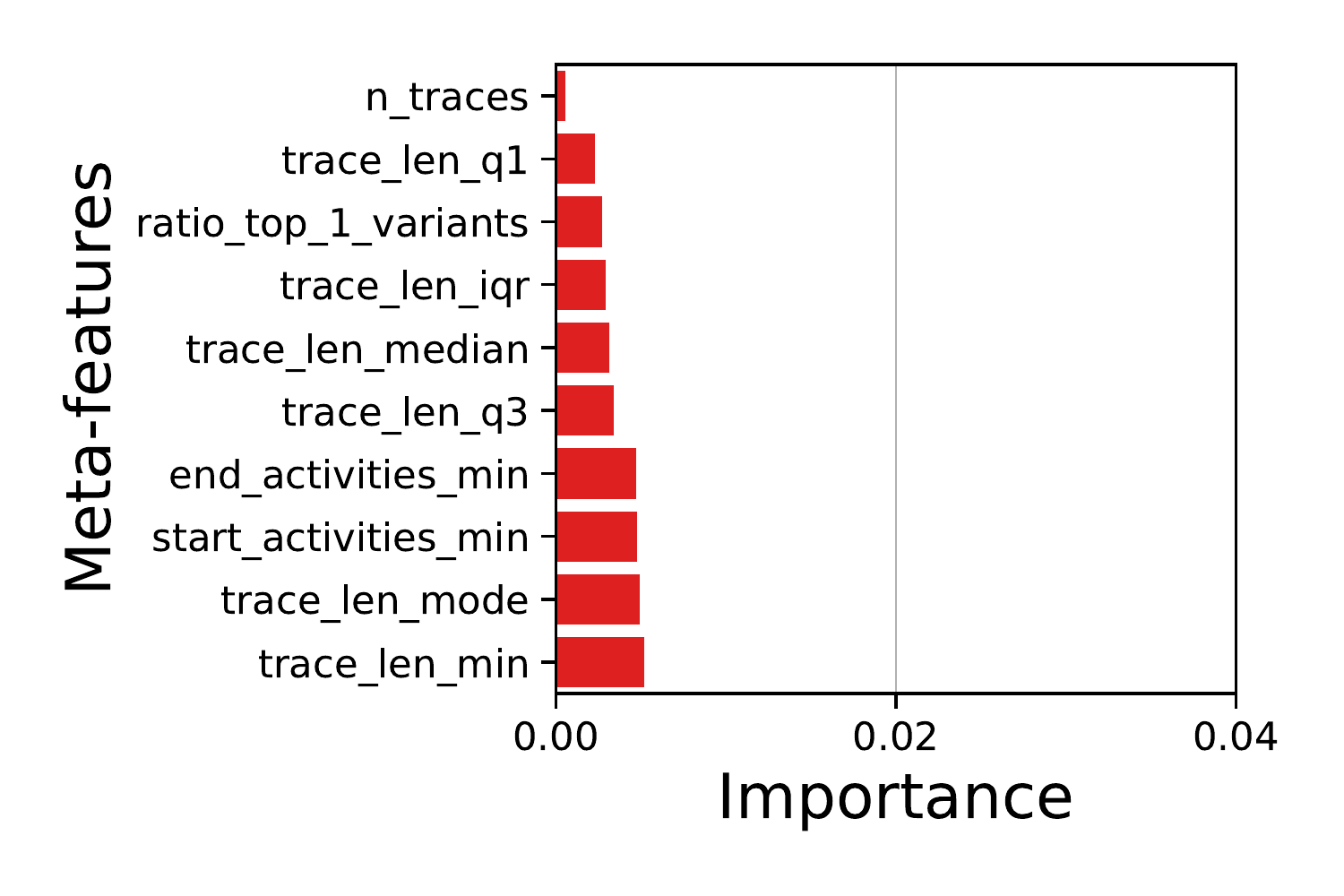}
        \caption{Bottom 10 features}
        \label{rank5_bot10_features}
    \end{subfigure}
    
    \caption{Features relevance to induce a meta-model recommending a discovery algorithm.}
\end{figure}

After the meta-model learning, the Random Forest provides an \textit{importance} measure to quantify the contribution of each feature provided to the classification output. Figures~\ref{rank5_top10_features} and~\ref{rank5_bot10_features} show that among the most influential features (Figure~\ref{rank5_top10_features}), there is a high predominance of the entropy family, namely, \textit{k}-block difference, Lempel-Ziv, trace, and \textit{k}-nearest neighbor. The entropy features were designed to capture log complexity and, more specifically, structuredness. Activities appear as the second most important group, with the 25th percentile ranking as the most important feature. The activity and entropy groups are correlated as both rely on activity information. Trace variants are the last group among the top 10 most influential features with some unique traces. These results highlight that high-level descriptors for event logs can aid the indication of the best discovery algorithm. Regarding the least influential features (Figure~\ref{rank5_bot10_features}), trace length-based features have the most appearances. The bottom position belongs to the number of traces. These results indicate that both the number of traces and trace lengths are the worst features to describe business process behavior.

\subsection{Stability of Experimental Design}

\begin{figure}[ht!]
    \centering
    \begin{subfigure}[b]{0.50\textwidth}    
       \includegraphics[width=1\textwidth]{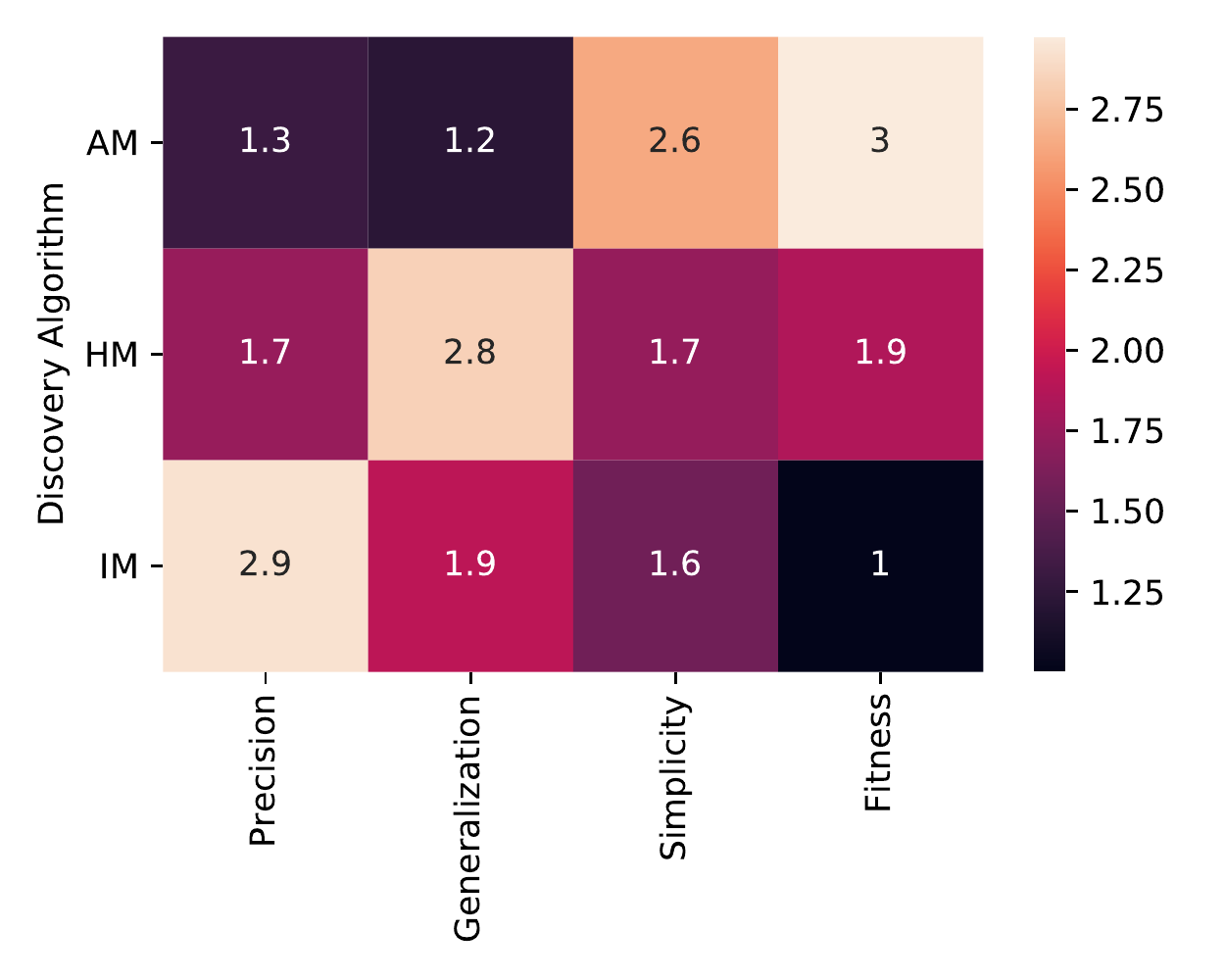}
    \caption{Metrics Ranking}
    \label{rank3}
    \end{subfigure}
    \hfill
    \begin{subfigure}[b]{0.49\textwidth}
        \includegraphics[width=1\textwidth]{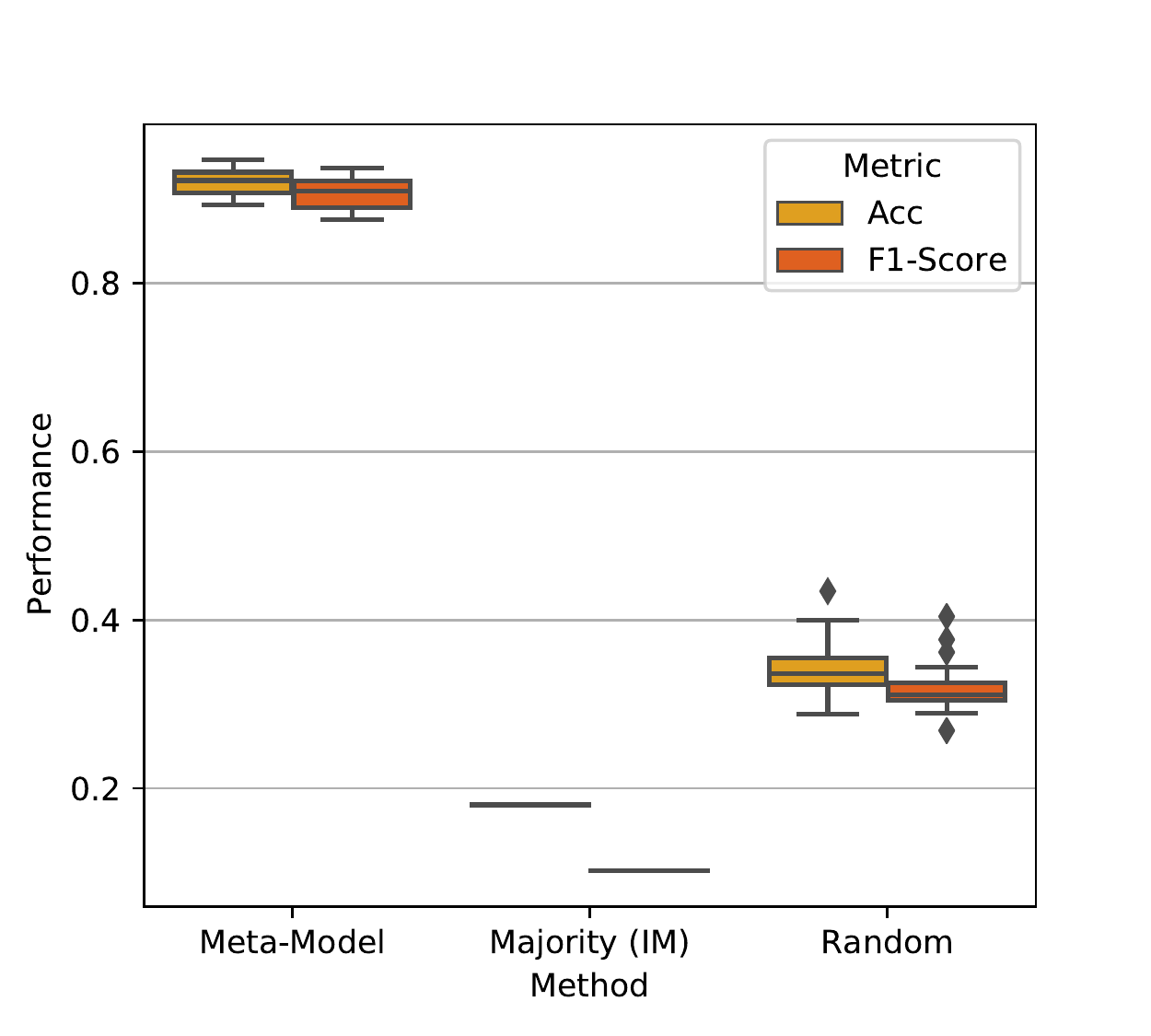}
    \caption{Recommendation performance}
    \label{rank3_perf}
    \end{subfigure}
    \caption{Discovery algorithm rankings across all event logs considering each dimension. Given event log features, the methods indicate the discovery algorithm that maximizes model quality. This experiment compares three discovery algorithms using four model quality criteria (fitness, precision, generalization and simplicity).}
\end{figure}

We repeated the experiment following the same standards but now considering only the four traditional quality metrics and three discovery techniques. In other words, we excluded the IM variants and the time perspective. Figure~\ref{rank3} shows the results of this experiment. Overall, the average positions relate to the previous analysis (Figure~\ref{rank5}). Regarding \textit{fitness}, IM maintains its position. HM comes next with a relatively higher average position (1.9) and AM averages at position 3, meaning that AM produces the lowest fitting model for all event logs. The same order is maintained in the simplicity analysis, but average positions are closer. IM, HM, and AM average at 1.6, 1.7, and 2.6, respectively. Producing simple models is not an easy task, especially when the meta-database contains many logs with anomalies, incompleteness, and complex behavior. AM ranks the best in \textit{precision} and \textit{generalization} (1.3 and 1.2), indicating its capability to generate a balanced model, neither underfitting nor overfitting. In precision, HM performs close to AM, averaging at 1.7, and is followed by IM averaging at 2.9. Again, AM and HM produce a model that balances well the escaping edges. IM's broad use of hidden transitions hurts its performance in this dimension. For generalization, second and third positions change, IM rises to the second-best algorithm (1.9) while HM falls to third (2.8). 

With fewer dimensions and algorithms, we can better evaluate some trade-offs offered by the discovery techniques. For instance, AM excels in precision and generalization, however, at the cost of having the worst simplicity and fitness performances. IM guarantees perfect fitness and often produces the simplest model. On the other hand, it has difficulties generating precise models and is only average in generalizing. HM is the most regular approach as it appears the second-best solution in three dimensions. However, it is never ranked as the best solution.

In this second experiment setup, the majority class was IM, i.e., the most recommended method across all event logs. Figure~\ref{rank3_perf} details the results for this experiment. The meta-model's performance reached even better results, with 0.92 of accuracy and 0.91 F-score, again confirming our approach's viability. The random approach followed with 0.34 and 0.32 of accuracy and F-score, respectively. Furthermore, with the worst performance, the majority method achieved an accuracy of 0.18 and an F-score of 0.1. These results show that the majority method has a worse efficiency when classes are more balanced.

\begin{figure}[ht!]
    \centering
    \begin{subfigure}[b]{0.49\textwidth}    
        \includegraphics[width=0.85\textwidth]{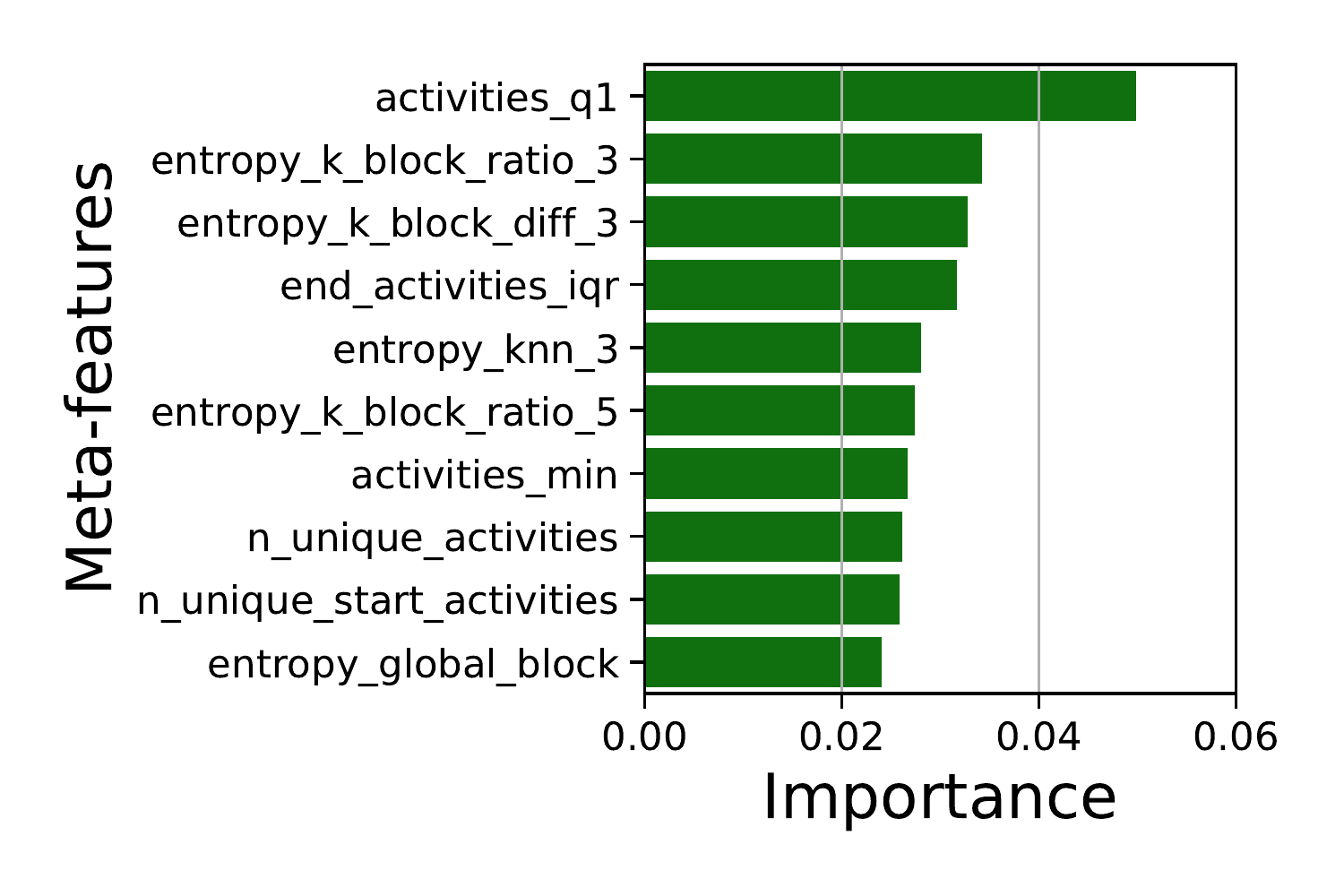}
        \caption{Top 10 features}
        \label{rank3_top10_features}
    \end{subfigure}
    \hfill
    \begin{subfigure}[b]{0.5\textwidth}
        \includegraphics[width=0.85\textwidth]{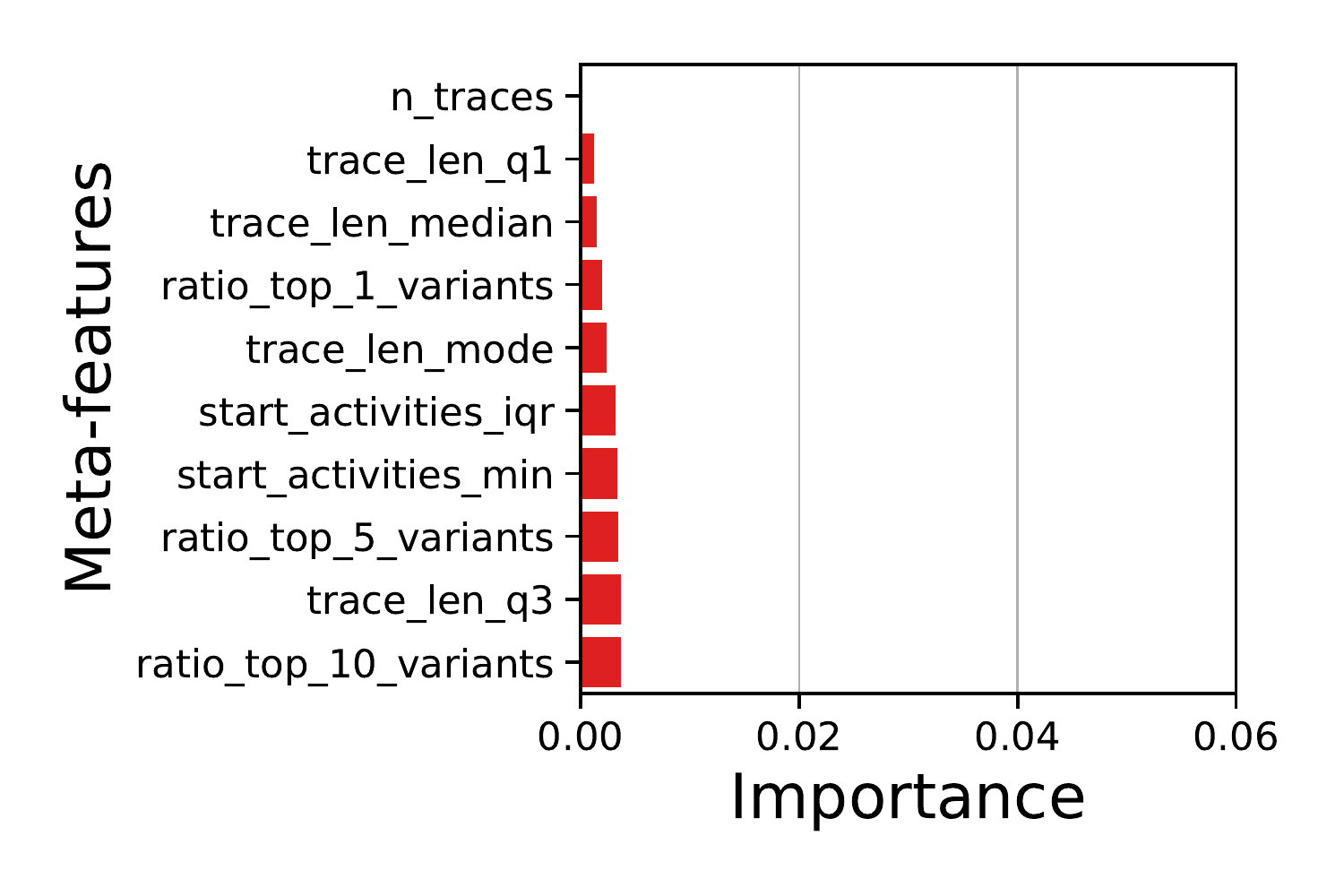}
        \caption{Bottom 10 features}
        \label{rank3_bot10_features}
    \end{subfigure}
    
    \caption{Features relevance to induce a meta-model recommending a discovery algorithm.}
\end{figure}

We also analyze the most (Figure~\ref{rank3_top10_features}) and less (Figure~\ref{rank3_bot10_features}) influential features in this scenario. As Figure~\ref{rank3_top10_features} shows, among the descriptors with the highest importance in the classification process, we observe the prevalence of the entropy family of features again, with five representatives in the top ten. The other five features belong to the activity family of features, showing the relevance of activity-related descriptors when evaluating an event log. Figure~\ref{rank3_bot10_features} confirms the low relevance of trace length-related features to capture process complexity.

\section{Conclusion}~\label{cc}
\vspace{-0.15em}

Discovering a process model for an event log might be a difficult task due to the many process discovery algorithms and characteristics of the log. This work proposes an MtL approach to map business process descriptors, discovery algorithms, and model quality criteria. We propose a set of 93 meta-features extracted from the event logs used as meta-instances in our experiment. The meta-features capture different process behavior at different levels, such as activity, trace, and log. Using an MtL, we show that it is possible to leverage process model quality by automatically recommending appropriate discovery algorithms. The best-fitting discovery method is the one that maximizes the model quality based on four traditional model metrics: fitness, precision, generalization, and simplicity. Our approach correctly assigns the best technique with 76\% and 92\% of accuracy with five and three discovery algorithms, respectively. Furthermore, the proposed MtL approach is highly extendable, enabling the possibility of encoding more features, discovery algorithms, and model metrics. As future work, we plan to increase the meta-database with more event logs, improve the set of meta-features and evaluate more specific trade-offs between model metrics. Also, we would like to cover other PM tasks using MtL approaches.

\bibliographystyle{splncs04}

\begin{thebibliography}{10}
\providecommand{\url}[1]{\texttt{#1}}
\providecommand{\urlprefix}{URL }
\providecommand{\doi}[1]{https://doi.org/#1}

\bibitem{aalstreplaying}
van~der Aalst, W., Adriansyah, A., van Dongen, B.: Replaying history on process
  models for conformance checking and performance analysis. WIREs Data Mining
  and Knowledge Discovery  \textbf{2}(2),  182--192 (2012)

\bibitem{Aalst16}
van~der Aalst, W.M.P.: Process Mining: Data Science in Action. Springer,
  Heidelberg, 2 edn. (2016)

\bibitem{van2010process}
Van~der Aalst, W.M., Rubin, V., Verbeek, H., van Dongen, B.F., Kindler, E.,
  G{\"u}nther, C.W.: Process mining: a two-step approach to balance between
  underfitting and overfitting. Software \& Systems Modeling  \textbf{9}(1),
  ~87 (2010)

\bibitem{aguiar2019meta}
Aguiar, G.J., Mantovani, R.G., Mastelini, S.M., de~Carvalho, A.C., Campos,
  G.F., Junior, S.B.: A meta-learning approach for selecting image segmentation
  algorithm. Pattern Recognition Letters  \textbf{128},  480--487 (2019)

\bibitem{Augusto2019TKDE}
{Augusto}, A., {Conforti}, R., {Dumas}, M., {Rosa}, M.L., {Maggi}, F.M.,
  {Marrella}, A., {Mecella}, M., {Soo}, A.: Automated discovery of process
  models from event logs: Review and benchmark. IEEE Transactions on Knowledge
  and Data Engineering  \textbf{31}(4),  686--705 (2019)

\bibitem{augusto2019split}
Augusto, A., Conforti, R., Dumas, M., La~Rosa, M., Polyvyanyy, A.: Split miner:
  automated discovery of accurate and simple business process models from event
  logs. Knowledge and Information Systems  \textbf{59}(2),  251--284 (2019)

\bibitem{Augusto2019}
Augusto, A., Dumas, M., La~Rosa, M.: Metaheuristic optimization for automated
  business process discovery. In: Hildebrandt, T., van Dongen, B.F.,
  R{\"o}glinger, M., Mendling, J. (eds.) Business Process Management. pp.
  268--285. Springer International Publishing, Cham (2019)

\bibitem{Back2019}
Back, C.O., Debois, S., Slaats, T.: Entropy as a measure of log variability.
  Journal on Data Semantics  \textbf{8}(2),  129--156 (Jun 2019)

\bibitem{10.1007/978-3-030-70650-0_11}
Barbon~Junior, S., Ceravolo, P., Damiani, E., Tavares, G.M.: Evaluating trace
  encoding methods in process mining. In: Bowles, J., Broccia, G., Nanni, M.
  (eds.) From Data to Models and Back. pp. 174--189. Springer International
  Publishing, Cham (2021)

\bibitem{BertiA19}
Berti, A., van~der Aalst, W.M.P.: Reviving token-based replay: Increasing speed
  while improving diagnostics. In: van~der Aalst, W.M.P., Bergenthum, R.,
  Carmona, J. (eds.) Proceedings of the International Workshop on Algorithms
  Theories for the Analysis of Event Data 2019 ATAED@Petri Nets/ACSD 2019.
  {CEUR} Workshop Proceedings, vol.~2371, pp. 87--103. CEUR-WS.org (2019)

\bibitem{breiman2001random}
Breiman, L.: Random forests. Machine learning  \textbf{45}(1),  5--32 (2001)

\bibitem{Buijs2014}
Buijs, J.C.A.M., van Dongen, B.F., van~der Aalst, W.M.P.: Quality dimensions in
  process discovery: The importance of fitness, precision, generalization and
  simplicity. International Journal of Cooperative Information Systems
  \textbf{23}(01) (2014)

\bibitem{9124702}
{Ceravolo}, P., {Tavares}, G.M., {Junior}, S.B., {Damiani}, E.: Evaluation
  goals for online process mining: a concept drift perspective. IEEE
  Transactions on Services Computing pp.~1--1 (2020)

\bibitem{DEWEERDT2012654}
{De Weerdt}, J., {De Backer}, M., Vanthienen, J., Baesens, B.: A
  multi-dimensional quality assessment of state-of-the-art process discovery
  algorithms using real-life event logs. Information Systems  \textbf{37}(7),
  654--676 (2012)

\bibitem{he2021automl}
He, X., Zhao, K., Chu, X.: Automl: A survey of the state-of-the-art.
  Knowledge-Based Systems  \textbf{212},  106622 (2021)

\bibitem{kerremans2018market}
Kerremans, M.: Market guide for process mining. Gartner Inc  (2018)

\bibitem{10.1007/978-3-642-38697-8_17}
Leemans, S.J.J., Fahland, D., van~der Aalst, W.M.P.: Discovering
  block-structured process models from event logs - a constructive approach.
  In: Colom, J.M., Desel, J. (eds.) Application and Theory of Petri Nets and
  Concurrency. pp. 311--329. Springer Berlin Heidelberg, Berlin, Heidelberg
  (2013)

\bibitem{10.1007/978-3-319-06257-0_6}
Leemans, S.J.J., Fahland, D., van~der Aalst, W.M.P.: Discovering
  block-structured process models from event logs containing infrequent
  behaviour. In: Lohmann, N., Song, M., Wohed, P. (eds.) Business Process
  Management Workshops. pp. 66--78. Springer International Publishing, Cham
  (2014)

\bibitem{10.1007/978-3-319-07734-5_6}
Leemans, S.J.J., Fahland, D., van~der Aalst, W.M.P.: Discovering
  block-structured process models from incomplete event logs. In: Ciardo, G.,
  Kindler, E. (eds.) Application and Theory of Petri Nets and Concurrency. pp.
  91--110. Springer International Publishing, Cham (2014)

\bibitem{10.1007/978-3-319-19237-6_6}
Leemans, S.J.J., Fahland, D., van~der Aalst, W.M.P.: Scalable process discovery
  with guarantees. In: Gaaloul, K., Schmidt, R., Nurcan, S., Guerreiro, S., Ma,
  Q. (eds.) Enterprise, Business-Process and Information Systems Modeling. pp.
  85--101. Springer International Publishing, Cham (2015)

\bibitem{muarucster2006rule}
M{\u{a}}ru{\c{s}}ter, L., Weijters, A.T., Van Der~Aalst, W.M., Van Den~Bosch,
  A.: A rule-based approach for process discovery: Dealing with noise and
  imbalance in process logs. Data mining and knowledge discovery
  \textbf{13}(1),  67--87 (2006)

\bibitem{de2007genetic}
de~Medeiros, A.K.A., Weijters, A.J., van~der Aalst, W.M.: Genetic process
  mining: an experimental evaluation. Data Mining and Knowledge Discovery
  \textbf{14}(2),  245--304 (2007)

\bibitem{mendling2008metrics}
Mendling, J.: Metrics for process models: empirical foundations of
  verification, error prediction, and guidelines for correctness, vol.~6.
  Springer Science \& Business Media (2008)

\bibitem{10.1007/978-3-642-15618-2_16}
Mu{\~{n}}oz-Gama, J., Carmona, J.: A fresh look at precision in process
  conformance. In: Hull, R., Mendling, J., Tai, S. (eds.) Business Process
  Management. pp. 211--226. Springer Berlin Heidelberg, Berlin, Heidelberg
  (2010)

\bibitem{oyamada2020towards}
Oyamada, R.S., Shimomura, L.C., Junior, S.B., Kaster, D.S.: Towards proximity
  graph auto-configuration: An approach based on meta-learning. In: European
  Conference on Advances in Databases and Information Systems. pp. 93--107.
  Springer (2020)

\bibitem{ROZINAT200864}
Rozinat, A., {van der Aalst}, W.: Conformance checking of processes based on
  monitoring real behavior. Information Systems  \textbf{33}(1),  64--95 (2008)

\bibitem{GARCIA2019260}
dos Santos~Garcia, C., Meincheim, A., {Faria Junior}, E.R., Dallagassa, M.R.,
  Sato, D.M.V., Carvalho, D.R., Santos, E.A.P., Scalabrin, E.E.: Process mining
  techniques and applications a systematic mapping study. Expert Systems with
  Applications  \textbf{133},  260--295 (2019)

\bibitem{1316839}
{van der Aalst}, W., {Weijters}, T., {Maruster}, L.: Workflow mining:
  discovering process models from event logs. IEEE Transactions on Knowledge
  and Data Engineering  \textbf{16}(9),  1128--1142 (2004)

\bibitem{VAZQUEZBARREIROS2015315}
Vázquez-Barreiros, B., Mucientes, M., Lama, M.: Prodigen: Mining complete,
  precise and minimal structure process models with a genetic algorithm.
  Information Sciences  \textbf{294},  315--333 (2015), innovative Applications
  of Artificial Neural Networks in Engineering

\bibitem{tonbeta166}
Weijters, A., Aalst, W., Medeiros, A.: {Process Mining with the Heuristics
  Miner-algorithm}. BETA Working Paper Series, WP 166, Eindhoven University of
  Technology, Eindhoven (2006)

\end{thebibliography}

\end{document}